 \documentclass[final,5p,times,twocolumn,authoryear]{elsarticle}

\usepackage{amssymb}
\usepackage{lipsum}
\usepackage{amsmath}
\usepackage{algorithm}
\usepackage{algorithmic}
\usepackage{subfigure} 

\journal{Elsevier}

\begin{document}

\begin{frontmatter}



\title{PriorFusion: Unified Integration of Priors for Robust Road Perception in Autonomous Driving}


\author[first]{Xuewei Tang}
\ead{tangxw20@mails.tsinghua.edu.cn}
\author[first]{Mengmeng Yang\corref{cor1}}
\author[first]{Tuopu Wen}
\author[first]{Peijin Jia}
\author[Second]{Le Cui}
\author[Second]{Mingshan Luo}
\author[Second]{Kehua Sheng}
\author[Second]{Bo Zhang}
\author[first]{Kun Jiang\corref{cor1}}
\author[first]{Diange Yang\corref{cor1}}

\cortext[cor1]{Corresponding author}


\affiliation[first]{organization={School of Vehicle and Mobility, Tsinghua University},
            addressline={}, 
            city={Beijing},
            postcode={100084}, 
            state={},
            country={China}}
\affiliation[second]{organization={Didi Chuxing},
            addressline={}, 
            city={Beijing},
            postcode={100084}, 
            state={},
            country={China}}
            
\begin{abstract}
With the growing interest in autonomous driving, there is an increasing demand for accurate and reliable road perception technologies. In complex environments without high-definition map support, autonomous vehicles must independently interpret their surroundings to ensure safe and robust decision-making. However, these scenarios pose significant challenges due to the large number, complex geometries, and frequent occlusions of road elements. A key limitation of existing approaches lies in their insufficient exploitation of the structured priors inherently present in road elements, resulting in irregular, inaccurate predictions.
To address this, we propose PriorFusion, a unified framework that effectively integrates semantic, geometric, and generative priors to enhance road element perception. We introduce an instance-aware attention mechanism guided by shape-prior features, then construct a data-driven shape template space that encodes low-dimensional representations of road elements, enabling clustering to generate anchor points as reference priors. We design a diffusion-based framework that leverages these prior anchors to generate accurate and complete predictions.
Experiments on large-scale autonomous driving datasets demonstrate that our method significantly improves perception accuracy, particularly under challenging conditions. Visualization results further confirm that our approach produces more accurate, regular, and coherent predictions of road elements.
\end{abstract}




\end{frontmatter}


\section{Introduction}
\label{introduction}

\begin{figure}[h]
    \centering
    \includegraphics[width=1\linewidth]{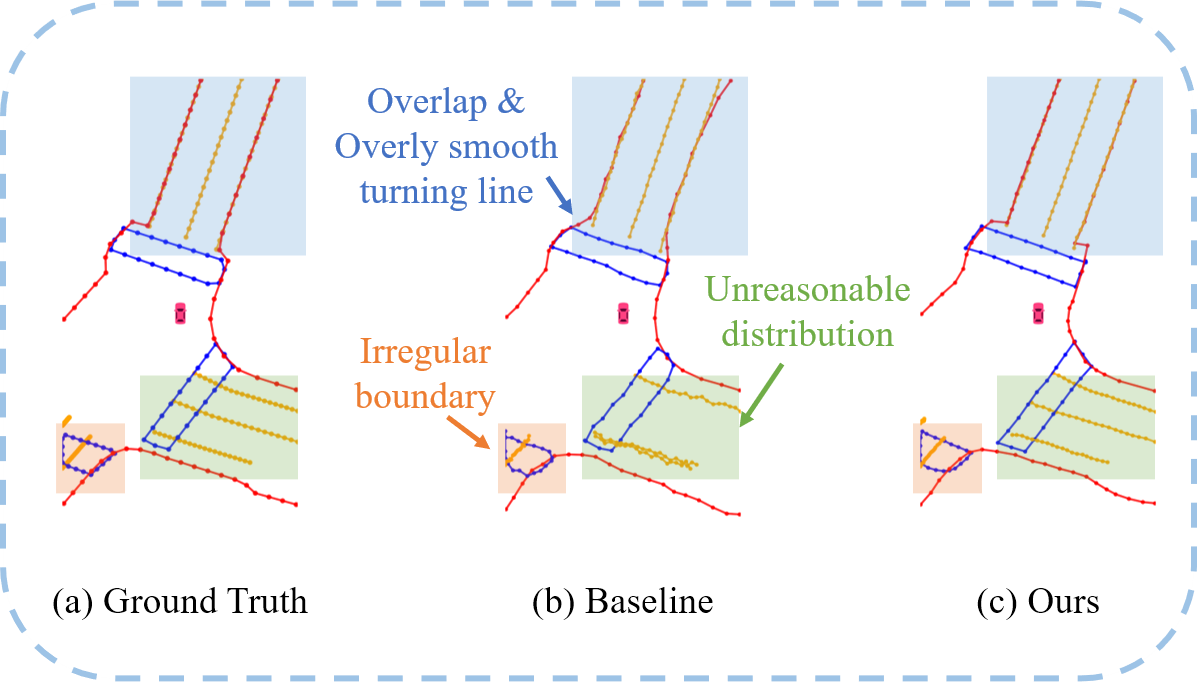}
    \caption{(a) shows the ground truth. (b) shows the problem of general methods lacking prior consideration of shape rules.(c) shows the prediction results of our model, which effectively makes the prediction distribution and shape more standardized and closer to the ground truth in (a).}
    \label{fig:problem}
\end{figure} 

High-Definition (HD) maps are crucial for safe navigation and decision making of autonomous vehicles, as they provide precise representations of static road elements such as lane markings, road boundaries, and pedestrian crossings \citep{yang2018intelligent,wen2022roadside,Seif2016autonomous,tang2023high}. However, the conventional HD map construction process is labor-intensive, costly, and infrequently updated, rendering it insufficient for dynamic and complex driving environments where rapid changes in road conditions and layouts occur \citep{sun2022gcdL,wijaya2022crowdsourced}. To address this limitation, there is a growing focus on integrating real-time road environment perception into onboard Bird’s Eye View (BEV) systems, enabling vehicles to dynamically infer road structures during driving \citep{peng2023bevsegformer, yang2023bevformerv2,  liu2022petrv2, huang2022bevpoolv2}.

Recent methods model road elements as objects to be detected in BEV space and achieve real-time perception through end-to-end network \citep{liu2022vectormapnet, liao2022maptr, yuan2023streammapnet}, where road elements are primarily represented as sets of vectorized polylines or polygons.

Despite their effectiveness in controlled scenarios, these methods cannot generalize well in complex environments (e.g. intersections, merges, and occluded areas) where road elements are abundant, structurally complex.

One of the key limitations of existing approaches lies in their limited use of shape prior knowledge. Road elements inherently exhibit geometric regularities, for example, lane lines tend to be elongated and continuous, pedestrian crossings are often generalized by parallelogram shapes, and road boundaries follow smooth curves. Neglecting these structural properties often leads to irregular, incomplete, and low accuracy predictions, which compromise the safety of downstream planning modules.
As shown in Fig.\ref{fig:problem} (b), the general method regards road elements as general objects perceived by BEV, while ignoring the shape and distribution rules of road elements, which may result in irregular boundaries, unreasonable element distribution, overly smooth shapes at turning points, overlapping and conflicting predictions of multiple elements. 
While several studies have attempted to incorporate prior knowledge into road perception, they underperform in complex scenarios due to incomplete prior utilization \citep{zhang2023onlinegemap} or heavy computational cost \citep{wang2024priormapnet}. 

In this paper, we explore the comprehensive utilization of the shape prior knowledge in road perception, and propose PriorFusion, a prior-enhanced perception framework designed to achieve accurate and consistent detection of road elements in complex scenarios. 
The main contributions of this work are summarized as follows:
 \begin{enumerate}

    \item Shape-Prior-Guided Query Refinement Module: 
    
    This module leverages shape priors from semantic segmentation to enhance the detection of vector instances.
    \item Shape Template Space Construction: 
    
    We build a low-dimensional shape template space and cluster annotations in this space to generate prior anchors that serve as reliable prior geometric references in map decoder.
    
    \item Shape-Prior-Enhanced Network with Diffusion Model: 
    
    We propose a novel diffusion-based architecture with low overhead that integrates prior anchors with the network backbone, enabling the generation of more complete and regular element shapes. 
    
\end{enumerate}
Finally, a series of experiments using state-of-the-art method as baseline demonstrate that our approach can effectively incorporate prior knowledge to improve prediction accuracy. Moreover, the proposed modules can be seamlessly integrated into existing map perception frameworks as \textbf{plug-and-play components} to enhance performance.
Visualization results show our method generates regular shapes, reduces boundary noise, and retains details at critical points. It also improves spatial distribution and reduces missed detections, enhancing overall prediction quality.
Notably, under stricter evaluation thresholds, our method achieves substantial performance gains, indicating that the predicted shapes align more closely with ground-truth annotations.
    
The paper is structured as follows: Section 2 reviews related methods in road element perception, including those using prior knowledge and shape rules, providing valuable insights. Section 3 analyses the shape prior mechanism of road elements and builds the module. Section 4 presents the experimental setup and results, and analyses the effectiveness of the model. In the last section, we give the conclusion and discuss some research challenges.

\section{Related Work}

High-definition (HD) maps are structured digital representations that provide comprehensive and accurate descriptions of road environments for autonomous vehicles. In scenarios where HD maps are unavailable or outdated, perception of road elements becomes essential for ensuring robust and safe end-to-end autonomous driving.
This section reviews existing research on road element perception and related detection methods, which form the foundation and provide key insights for the proposed approach.
 
\subsection{Vectorized Road Elements Perception}
Since HDMapNet \citep{li2022hdmapnet} models local road elements as learnable objects for neural networks, researches have focused on utilizing single-network frameworks for road elements perception.VectorMapNet \citep{liu2023vectormapnet} was the first to represent road elements as vectors, enabling the end-to-end detection. Building upon this, MapTR \citep{liao2022maptr} integrates task of road element detecion within the object detection framework of DETR \citep{DETR}, representing road elements as fixed-length vector point chains. It introduces a unique representation of road elements by designing a set of uniquely defined points and expressions to eliminate ambiguity. Additionally, MapTR \citep{liao2022maptr} employs a hierarchical query structure that models road elements through instance-level and point-level queries, allowing the neural network to consistently output structured vectorized road elements.
Subsequent advancements, including MapTRv2 \citep{liao2023maptrv2}, InsightMapper \citep{xu2023insightmapper}, MapQR \citep{liu2024mapqr}, and HIMap \citep{zhou2024himap}, have further refined this approach. These works introduced novel query mechanisms and attention mechanisms to improve the accuracy of road element detection. 

However, these methodologies predominantly treat road elements as general objects of BEV object detection, often overlooking the inherent regularities in the shape and distribution of road elements. This limitation can lead to unstable shape predictions, such as irregular representations of sidewalks and road edges. Such inaccuracies not only reduce the overall detection precision but also risk introducing confusion in downstream modules that rely on the road elements for further processing.
\begin{figure*}[h]
    \centering
    \includegraphics[width=0.9\linewidth]{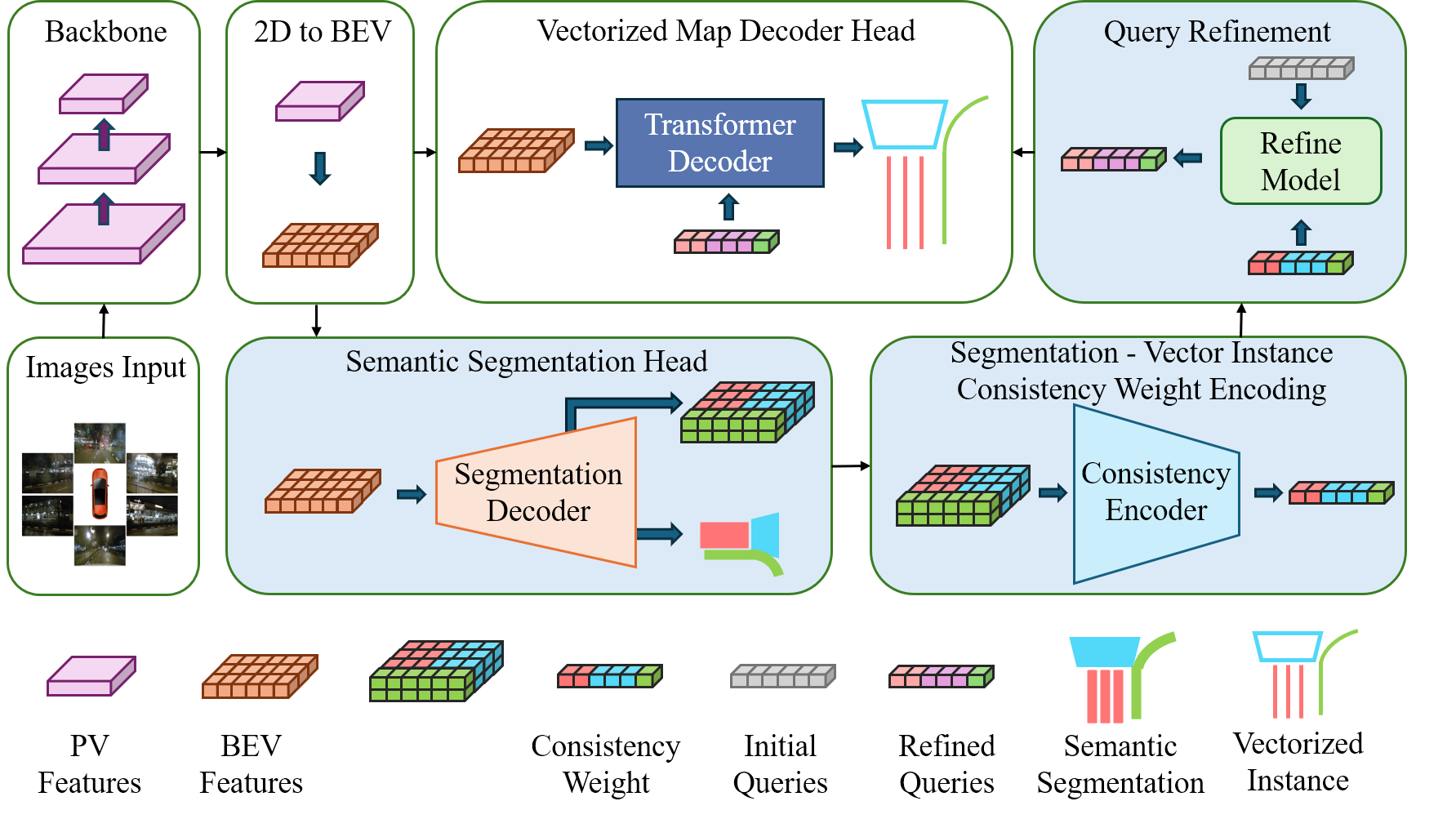}
    \caption{The pipeline of our vectorized road element detection model.}
    \label{fig:segpipeline}
\end{figure*}

\subsection{Enhancing Road Elements Perception Performance with Prior}
Road elements, as structured data models, have rich prior information compared to the diverse learning objects encountered in general vision tasks. The process of drawing road elements follows well-defined arrangement rules and standardised shape criteria. These rules provide structured knowledge that is valuable for road element detecion.

GeMap \citep{zhang2023onlinegemap} was one of the first methods to consider the integration of shape rules by modeling topological relationships and introducing a topology-geometry decoupling attention mechanism. It also employed a Euclidean spatial shape loss to facilitate network convergence.
However, due to the limited exploitation of prior information, its visual results still exhibit irregular and jagged shape boundaries, indicating suboptimal geometric consistency.
MapPrior \citep{zhu2023mapprior} and DiffMap \citep{jia2024diffmap} leverage diffusion models to learn prior information and generate high-quality semantic segmentation of road scenes. However, these methods are specifically designed for segmentation tasks and lack the capability to produce vectorized road element instances.
Moreover, diffusion-based approaches are computationally intensive, which severely limits their suitability for real-time onboard perception. For instance, MapPrior \citep{zhu2023mapprior} achieves an inference speed of only 0.60 FPS, while DiffMap \citep{jia2024diffmap} operates at just 0.32 FPS, making them impractical for latency-sensitive autonomous driving applications.

PriorMapNet \citep{wang2024priormapnet} performs offline clustering of ground-truth labels to generate initial reference points for the decoder structure. However, clustering in high-dimensional space on large-scale datasets can lead to excessive computational resource consumption. Therefore, it is essential to explore dimensionality reduction techniques to cope with the growing volume of autonomous driving data and the increasing diversity of driving scenarios.

Although prior knowledge has been increasingly applied to road element extraction, existing methods often underutilize such information and rarely conduct a comprehensive analysis of priors to guide model design. As a result, the predicted shapes and spatial layouts of road elements frequently deviate from expected structural patterns, leading to reduced perception accuracy in challenging environments.

\subsection{Integrating Shape Prior for Detection and Prediction}

SVD and principal component analysis (PCA) are widely used for compact data representation across various domains, such as Eigenfaces for face recognition  \citep{turk1991eigenfaces}. The eigenspace obtained through SVD decomposition captures essential features for reconstructing shapes. For example, Eigencontours \citep{park2022eigencontours} describe object boundaries using SVD-based data-driven shape descriptors, enabling faithful object representation with minimal coefficients.

Building on this concept, Eigenlanes \citep{jin2022eigenlanes} leverage SVD to represent road lanes by learning their distribution in training data, avoiding parametric curves like polynomials \citep{neven2018towards} or splines \citep{bezierlanedetection}. Similarly, EigenTrajectory \citep{bae2023eigentrajectory} constructs an eigenspace to express trajectories and generate trajectory anchors. These methods use pre-trained principal components to capture shape characteristics, representing shapes as linear combinations of a small number of bases for precise and efficient prediction. This approach leverages data-driven pretraining to effectively extract robust shape descriptors.

The diffusion model has proven to be an effective method for leveraging prior knowledge. However, existing applications of diffusion models in semantic map segmentation, such as MapPrior and DiffMap \citep{zhu2023mapprior, jia2024diffmap}, are hindered by their time-consuming nature and high computational demands. 
DiffusionDrive \citep{diffusiondrive} introduced a truncated diffusion model, which enables the prediction of more accurate vectorized trajectories within a limited number of steps. This innovation suggests that implementing diffusion models with restricted steps could significantly enhance the efficiency of online mapping.

\section{Methodology}

\subsection{Shape-Prior-Guided Query Refinement Module}
The representation of road elements primarily takes two forms: semantic segmentation masks and vectorized instances. These forms describe the same elements but in different data formats. Ideally, the keypoints in vectorized instances should match the shape and spatial layout of the corresponding semantic segmentation results.
Leveraging this relationship, using semantic segmentation masks as prior knowledge to guide the regression of keypoints for vectorized instances can ensure structural consistency between the two representations. This enhances the stability of keypoint coordinate regression.
Some existing methods explore using supervised semantic segmentation results to improve the quality of BEV features. However, this supervision is too deep and needs a method to link features close to the final semantic segmentation with the vector instance decoder.

Building on this concept, this study introduces a Shape-Prior-Guided Query Enhancement Network, as shown in Fig. \ref{fig:segpipeline}. 
Specifically, the model extracts semantic segmentation features from the layer preceding the final segmentation output, preserving rich shape information. These features are then processed through a feature transformation module to generate weight values that encode segmentation prior. The query enhancement module applies these segmentation weight values to the detection queries, effectively embedding semantic segmentation information into the query representations. This integration allows the detection queries to inherit shape-aware prior knowledge, thereby enhancing the accuracy and stability of road element perception.

Then, we specifically introduce the innovation module.

\subsubsection{Segmentation-Vector Instance Consistency Weight Encoding}

\begin{figure*}[h]
    \centering
    \includegraphics[width=0.7\linewidth]{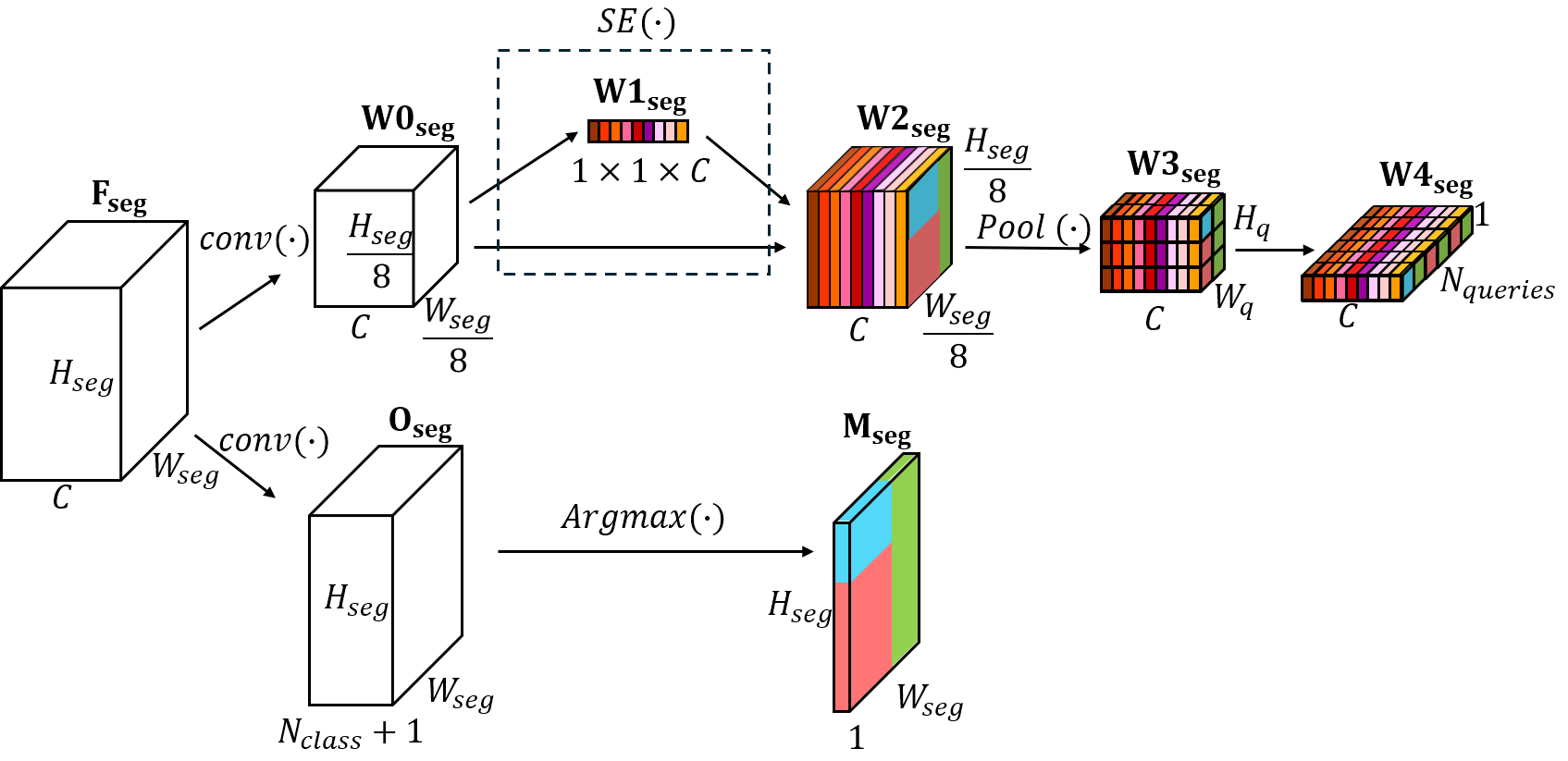}
    \caption{The Segmentation-Vector instance consistency weight encoding module.}
    \label{fig:semanticprior}
\end{figure*} 
The Segmentation-Vector Instance Consistency Weight Encoding module in Fig.\ref{fig:segpipeline} is detailed in Fig.\ref{fig:semanticprior}.

After the BEV features $\mathbf{F}_{\text{BEV}} \in \mathbb{R}^{H_{BEV} \times W_{BEV} \times C}$ are obtained through feature extraction networks, $\mathbf{F}_{\text{BEV}}$ are fed into a semantic segmentation decoder $D_{\text{Seglayer1}}$ consisting of an up-sampling module to get the intermediate layer semantic segmentation features $\mathbf{F}_{\text{Seg}}$.

\begin{equation}
  \mathbf{F}_{\text{Seg}} = D_{\text{Seglayer1}} (\mathbf{F}_{\text{BEV}})
\label{eq:chap03-eq01}
\end{equation}

$\mathbf{F}_{\text{Seg}}$ have $C$ channels and match the segmentation mask in size, $\mathbf{F}_{\text{Seg}} \in \mathbb{R}^{H_{seg} \times W_{seg} \times C}$. Then, $\mathbf{F}_{\text{Seg}}$ are fed into the next decoder $D_{\text{Seglayer2}}$ to get the semantic segmentation output $\mathbf{O}_{\text{Seg}} \in \mathbb{R}^{H_{seg} \times W_{seg} \times (N_{class} + 1)}$.
Through Argmax function, we get the final segmentation mask $\mathbf{M}_{\text{Seg}}$.

\begin{equation}
  \mathbf{M}_{\text{Seg}} = \text{Argmax} \left( D_{\text{Seglayer2}} \left( D_{\text{Seglayer1}} (\mathbf{F}_{\text{BEV}}) \right) \right)
  \label{eq:chap03-eq04}
\end{equation}

By supervising $\mathbf{M}_{\text{Seg}}$, we can obtain features $\mathbf{F}_{\text{Seg}}$ closest to the semantic segmentation output $\mathbf{O}_{\text{Seg}}$.
Then, $\mathbf{F}_{\text{Seg}}$ are fed into a size alignment module to reduce feature size while maintaining the number of channels, ultimately adapting them to the number of queries.

To achieve this, the feature size is quickly reduced from $H_{seg} \times W_{seg} \times C$ to $\frac{H_{seg}}{8} \times \frac{W_{seg}}{8} \times C$ through a size reduction module $\text{Module}{\text{conv}}$ composed of several convolutional layers, thereby decreasing the computational burden of feature weight conversion. The initial weights $\mathbf{W0}_{\text{Seg}}$, where $\mathbf{W0}_{\text{Seg}} \in \mathbb{R}^{\frac{H{seg}}{8} \times \frac{W_{seg}}{8} \times C}$, are obtained as follows:

\begin{equation}
  \mathbf{W0}_{\text{Seg}} = \text{Module}_{\text{conv}}(\mathbf{F}_{\text{Seg}})
  \label{eq:chap03-eq05}
\end{equation}

Next, an SE module  \citep{hu2018squeeze} is introduced to perform channel-wise attention calculation. This process follows the standard SE module procedure, where global average pooling, squeeze, and excitation operations are used to compute the channel-wise weights $\mathbf{W1}_{\text{Seg}} \in \mathbb{R}^{1 \times 1 \times C}$:
\begin{equation}
  \mathbf{W1}_{\text{Seg}} = \text{SE}(\mathbf{W0}_{\text{Seg}})
  \label{eq:chap03-eq06}
\end{equation}

Element-wise multiplication is then applied across the channel dimension using the broadcast mechanism, producing the weight $\mathbf{W2}_{\text{Seg}}$, where $\mathbf{W2}_{\text{Seg}} \in \mathbb{R}^{\frac{H_{\text{seg}}}{8} \times \frac{W_{\text{seg}}}{8} \times C}$. This operation retains the information in $\mathbf{W0}_{\text{Seg}}$ while minimizing information loss.

\begin{equation}
  \mathbf{W2}_{\text{Seg}} = \mathbf{W1}_{\text{Seg}} \odot \mathbf{W0}_{\text{Seg}}
  \label{eq:chap03-eq07}
\end{equation}

However, there is still a discrepancy between the size $\frac{H_{\text{seg}}}{8} \times \frac{W_{\text{seg}}}{8} \times C$ and the target size of $N_{queries} \times C$. To align with the query number, adaptive average pooling is applied, which effectively preserves feature map information compared to max pooling, yielding $\mathbf{W3}_{\text{Seg}} \in \mathbb{R}^{H_q \times W_q \times C}$:

\begin{equation}
  \mathbf{W3}_{\text{Seg}} = \text{AdaptiveAvgPool}(\mathbf{W2}_{\text{Seg}})
  \label{eq:chap03-eq08}
\end{equation}
where $N_{\text{queries}} = H_q \times W_q$.

Finally, $\mathbf{W3}_{\text{Seg}}$ is flattened along the width dimension to obtain the final query weights $\mathbf{W4}_{\text{Seg}} \in \mathbb{R}^{N_{\text{queries}}\times C}$.

\subsubsection{Query Refinement}
To assign the query weights containing semantic segmentation information $\mathbf{W4}_{\text{Seg}} \in \mathbb{R}^{N{\text{queries}} \times C}$ to the detection queries $\mathbf{Q}_{detection} \in \mathbb{R}^{N{\text{queries}} \times C}$, a query refinement module is used for integration. This process can be expressed as:
\begin{equation}
  \mathbf{Q}_{refined} = Module_{refine}(\mathbf{W4}_{\text{Seg}} , \mathbf{Q}_{detection})
  \label{eq:chap03-eq11}
\end{equation}

Various information fusion modules can be applied in this process. For example, a multi-layer perceptron (MLP) can be used, where the weights are combined with the queries as a matrix, and the MLP establishes an information connection between the weights and queries, ultimately producing the refined query $\mathbf{Q}_{refined} \in \mathbb{R}^{N{\text{queries}} \times C}$.

To simplify computation and reduce the module's computational cost, this model directly applies element-wise multiplication for weight assignment:
\begin{equation}
  \mathbf{Q}_{refined} = \mathbf{W4}_{\text{Seg}} \odot \mathbf{Q}_{detection}
  \label{eq:chap03-eq12}
\end{equation}

\subsection{Data-Driven Shape Template Space Construction and Prior Anchor Design}

\subsubsection{Shape Template Space Construction}
A road element can be described as a set of two-dimensional keypoints with a connected sequence, represented by the horizontal and vertical coordinates of their own vehicle coordinate system.
\begin{equation}
  \mathbf{r} = \begin{bmatrix} x_1, y_1, x_2, y_2, \dots, x_P, y_P \end{bmatrix}^\top
\end{equation}
where $x_i$ is the $x$-coordinate of the $i$-th sample and $P$ is the fixed number of keypoints. 
Each road element is composed of $N=2P$ numerical values. All road elements in the dataset can be combined into the road element matrix $A$ of the dataset:

\begin{equation}
  \mathbf{A} = \begin{bmatrix} \mathbf{r}_1, \mathbf{r}_2, \dots, \mathbf{r}_L \end{bmatrix}^\top
\end{equation}
where $L$ denotes the total number of road element instances.
Then, we apply SVD to the road element matrix $A$ as follows:

\begin{equation}
  \mathbf{A} = \mathbf{U} \boldsymbol{\Sigma} \mathbf{V}^\top
\end{equation}

where $\mathbf{U} = [\mathbf{u}_1, \dots, \mathbf{u}_N]$ and $\mathbf{V} = [\mathbf{v}_1, \dots, \mathbf{v}_L]$ are orthogonal matrices and $\boldsymbol{\Sigma}$ is a diagonal matrix composed of singular values $\sigma_1 \geq \sigma_2 \geq \cdots \geq \sigma_r > 0$.
Here, $r$ is the rank of road element matrix $A$.

Therefore, the best M rank approximation of A is:

\begin{equation}
  \mathbf{A}_M = \begin{bmatrix} \tilde{r}_1, \dots, \tilde{r}_L \end{bmatrix} = \sigma_1 \mathbf{u}_1 \mathbf{v}_1^\top + \cdots + \sigma_M \mathbf{u}_M \mathbf{v}_M^\top
  \label{eq:matrix12}
\end{equation}
in which the Frobenius norm $\|\mathbf{A} - \mathbf{A}_M\|_F$ is minimized.

The sum of squared errors for approximation of road elements is:
\begin{equation}
  \|\mathbf{A} - \mathbf{A}_M\|_F^2 = \sum_{i=1}^{L} \|\mathbf{x}_i - \tilde{\mathbf{x}}_i\|^2 = \sum_{i=M+1}^{r} \sigma_i^2
  \label{eq:matrix13}
\end{equation}

In equation \ref{eq:matrixa}, the estimation of each road element can be represented by a linear combination of the first M left singular vectors  $\mathbf{u}_1, \dots, \mathbf{u}_M$ of matrix $\mathbf{U}$:

\begin{equation}
  \tilde{\mathbf{r}}_i = \mathbf{U}_M \mathbf{c}_i = \begin{bmatrix} \mathbf{u}_1, \dots, \mathbf{u}_M \end{bmatrix} \mathbf{c}_i
\end{equation}

Then, we take these M vectors $\mathbf{u}_1, \dots, \mathbf{u}_M$ of matrix $\mathbf{U}$ as the basis of the shape space of road elements, which is the shape template and also the first M principal components of principal component analysis.
We span these M vectors into a shape template space, because they are eigenvectors of $AA^\top$. Therefore, as long as there is any road element $\mathbf{r}$, we can map it to the shape template space to obtain an approximate value of the road element:

\begin{equation}
  \widetilde{\mathbf{r}} = \mathbf{U}_M \mathbf{c}
  \label{eq:chap03-x-estimate}
\end{equation}
in which the coefficient c is obtained by the following equation:
\begin{equation}
  \mathbf{c} = \mathbf{U}_M^\top \mathbf{r}
  \label{eq:chap03-c-estimate}
\end{equation}

\begin{figure}[htbp]
  \centering
    \subfigure[All elements.]{\includegraphics[width=0.23\textwidth]{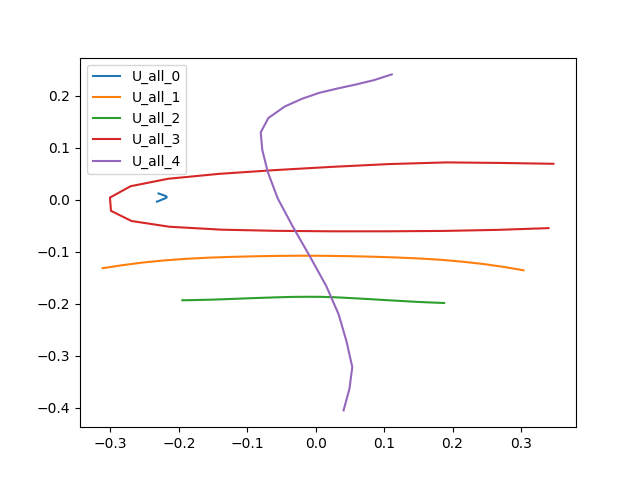}} 
    \subfigure[Pedestrian crossings.]{\includegraphics[width=0.23\textwidth]{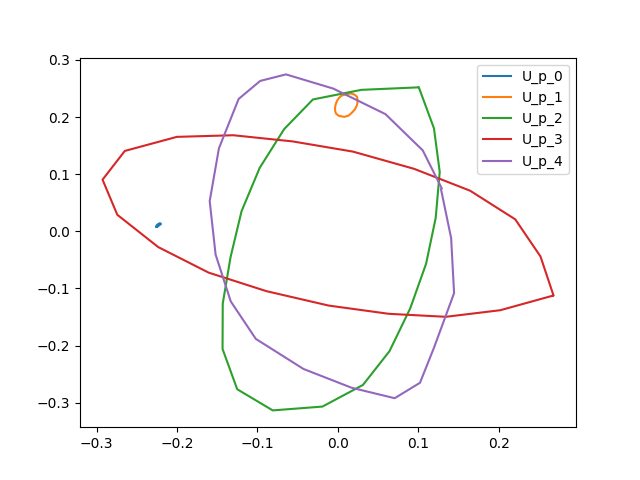}} 
    \vfill
    \subfigure[Dividers.]{\includegraphics[width=0.23\textwidth]{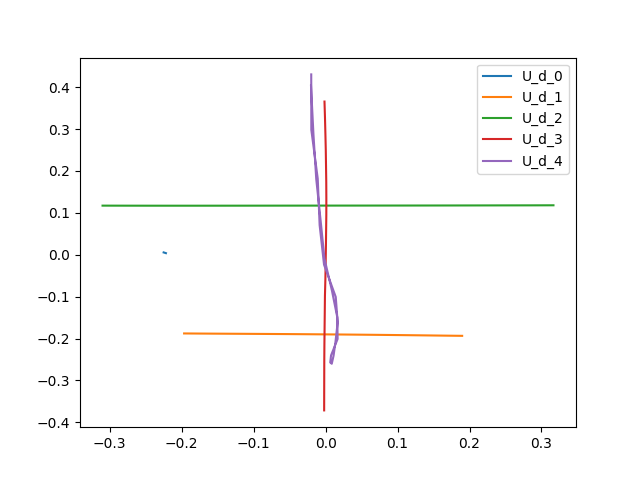}} 
    \subfigure[Boundaries.]{\includegraphics[width=0.23\textwidth]{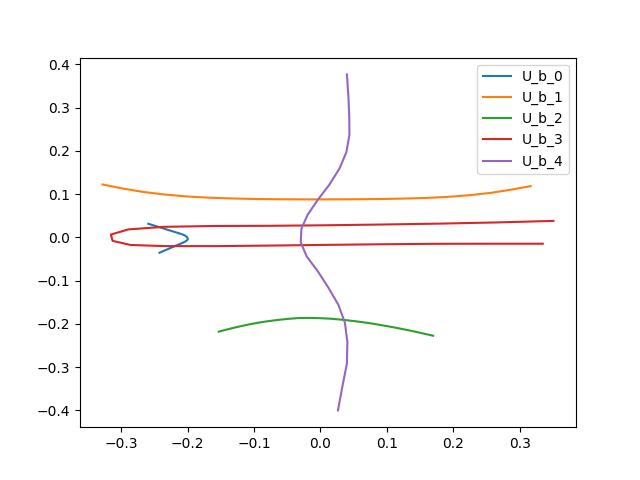}}

  \caption{Visualization of the first 5 eigenvectors for all road elements, pedestrian crossings, dividers, and boundaries.}
  \label{fig:template_vis}
  \vspace{0.2in}
\end{figure}

Therefore, in the shape template space, a road element $\mathbf{r}$ can be represented by M N-dimensional vectors $\mathbf{c}$ using the equation \ref{eq:chap03-x-estimate}.
Then we can transform the learning task of N coordinate values of road elements into a regression task of M template vectors, which are pre-trained using the dataset ground truth values. The visualization of the first five eigenvectors of the data matrix, comprising all road elements, pedestrian crossings, dividers, and boundaries, is presented in Fig. \ref{fig:template_vis}. The visualization results demonstrate that eigenvectors effectively capture the shape characteristics of different road elements. Specifically, the eigenvectors representing all elements and boundary elements exhibit higher similarity, likely because boundaries are more frequently encountered and have a more complex shape distribution than dividers. 

Furthermore, the templates, derived from a large dataset, exhibit rotation and translation invariance, making them robust to variations in vehicle orientation. This is particularly evident in the pedestrian crossing templates, where components are displayed in various orientations.

\subsubsection{Selection of Prior Anchors in Shape Template Space}

\begin{figure*}[p]
  \centering
    \subfigure[Reference points obtained through random initialization.]{\includegraphics[width=0.45\textwidth]{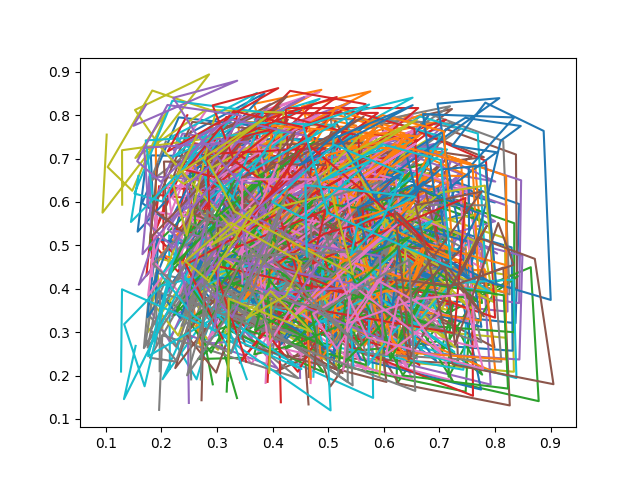}} 
    \subfigure[Prior reference points obtained through clustering in the template space.]{\includegraphics[width=0.45\textwidth]{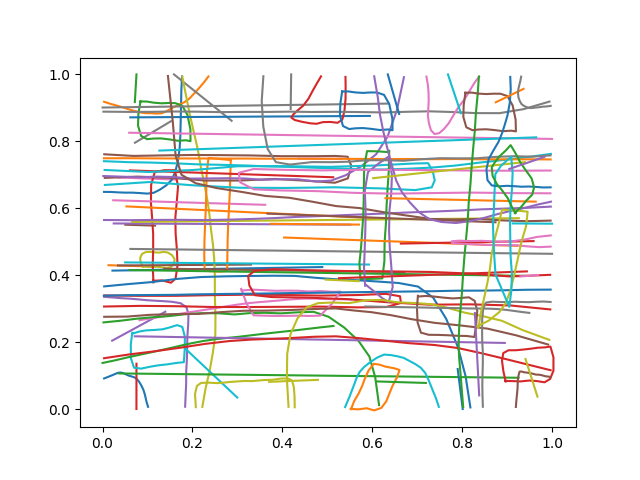}} 

  \caption{Compare the visualization of prior reference points obtained through random initialization and clustering in template space. Randomly initialized points are scattered across the entire region of interest without any specific pattern. In contrast, points derived from annotated data exhibit clear distribution patterns and distinct types, reflecting their basis in real-world annotations.}
  \label{fig:anchor_vis}
  \vspace{0.2in}
\end{figure*}

To capture typical spatial patterns and shape characteristics of road environment elements, a clustering algorithm can identify the most frequent shapes in the dataset  \citep{wang2024priormapnet}. However, K-Means, the most common method, is computationally expensive and prone to memory issues as the dataset size increases.

After constructing the shape template space, this study effectively maps the dataset’s element set from a Euclidean coordinate space matrix $\mathbf{A}$ of size $N \times L$ to a shape template space coefficient matrix $\mathbf{C}_A$ of size $M \times L$ using the shape template matrix $\mathbf{U}_M$:
\begin{equation}
  \mathbf{C}_A = \mathbf{U}_M^\top \mathbf{A}
  \label{eq:chap03-eq21}
\end{equation}
Since the shape template matrix $\mathbf{U}_M$ is an orthogonal matrix, it preserves the Euclidean distances between vectors, making the mapping between Euclidean space and shape template space isometric. Clustering in the shape template space maintains the geometric relationships between data points, making it equivalent to direct clustering in Euclidean space while significantly reducing the parameter count in the clustering process.

Therefore, despite road elements having both linear and nonlinear shapes, the clustering process is equivalent due to the isometry between the template space and the vector sampling point space. Moreover, as shown in equation \ref{eq:matrix12} and \ref{eq:matrix13}, there is some error when reconstructing the original space from the template space, which is determined by the discarded $N-M$ dimensional coefficients. According to the relationship of PCA principal component coefficients, the later principal components have smaller coefficients, so this error is tolerable.

Consequently, the clustering process for the dataset element matrix $\mathbf{A}$ is transformed into clustering the shape template space coefficient matrix $\mathbf{C}_A$.

Algorithm\ref{alg:kmeans} demonstrates the process of clustering element data of dataset in shape template space to obtain the final priority anchor. 
In this algorithm, 
$L$ is the number of all elements in the dataset,  $N_P$ is the number of prior anchors, for this model, we set to the number of queries $N_{queries}$, the number of maximum iterations $S_{\text{max}}$ is set to 300, the convergence threshold $\epsilon_{A}$ is set to $1e^{-4}$.

By preprocessing the dataset element matrix $\mathbf{A}$ using this algorithm, $N_P$ typical road environment element prior coordinates $P_A=\{{anchor}_{k}\}^{N_P}_{k=1}$ are obtained. The prior values obtained through clustering capture the typical shape and spatial distribution of map elements, providing reliable initial reference points for keypoint coordinate regression in the neural network. Unlike randomly initialized points, these clustered priors leverage shape information, leading to faster and more stable network convergence. As shown in Fig.\ref{fig:anchor_vis}, the clustered reference points more accurately reflect the shape and distribution of road elements compared to randomly initialized ones.

\begin{figure*}[htbp]
  \centering
    \subfigure[MapQR Decoder]{\includegraphics[width=0.45\textwidth]{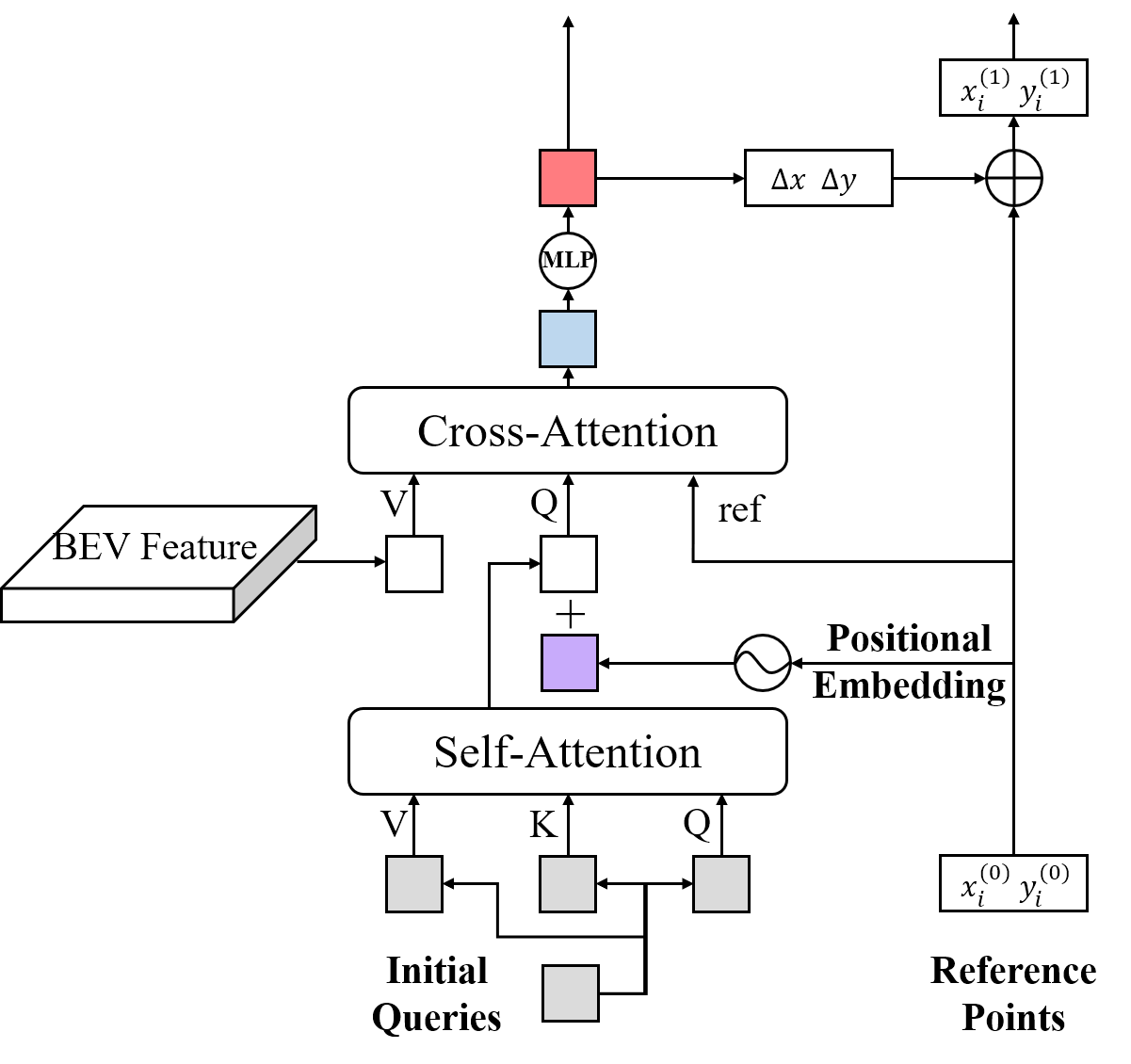}} 
    \subfigure[Ours PriorFusion Decoder]{\includegraphics[width=0.45\textwidth]{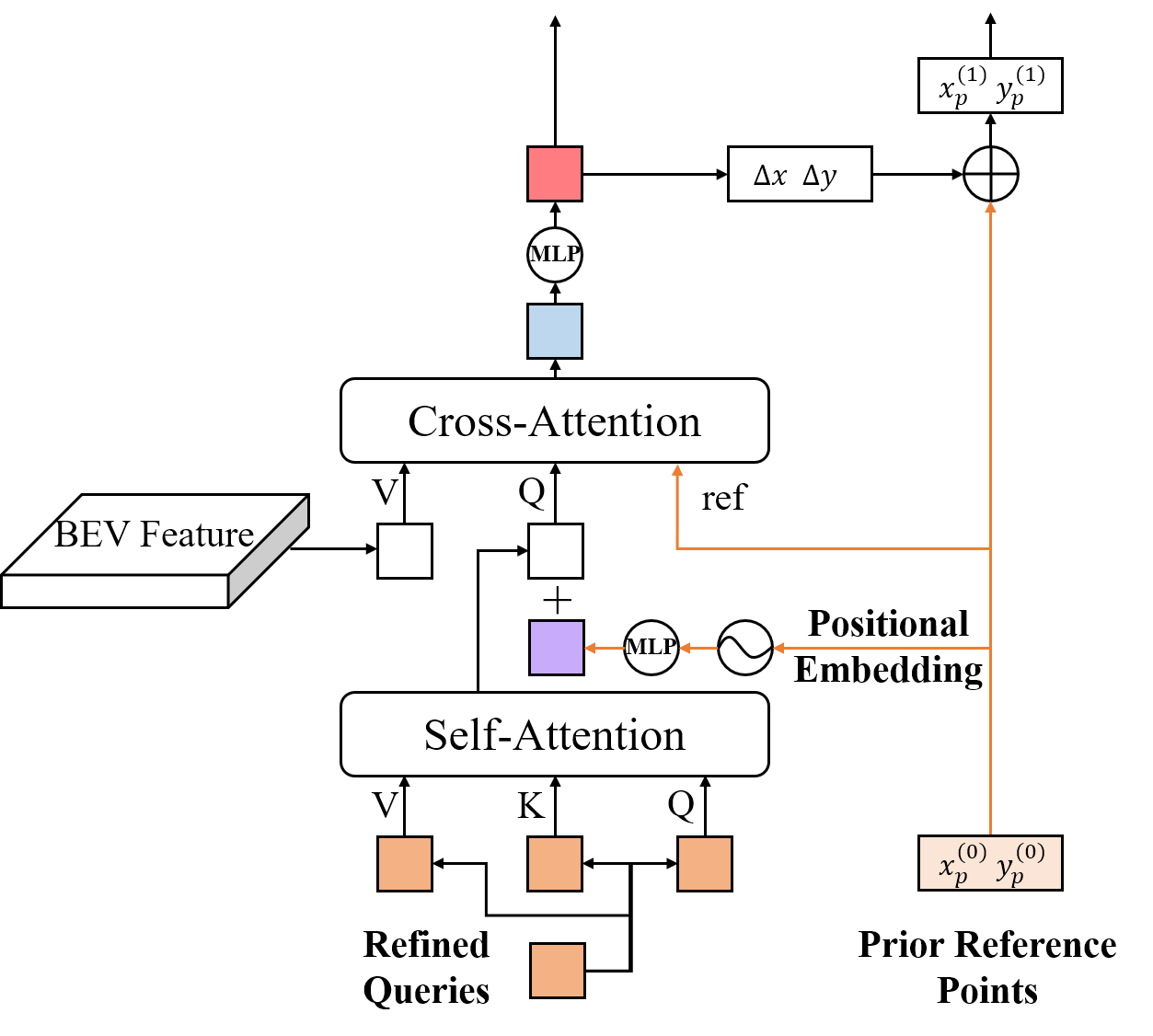}} 

  \caption{Compare the decoder of baseline model and our method.}
  \label{fig:mapqr_and_ours}
  \vspace{0.2in}
\end{figure*}

To summarize the above model, we compare the decoder with the baseline model decoder as shown in Fig.\ref{fig:mapqr_and_ours}. 
In this figure, we illustrate the decoder of the baseline model and the decoder of our PriorFusion. 
We replace the initial queries with the refined queries obtained from the aforementioned SPG model to acquire attention information on shapes from the semantic segmentation priors of the current frame. 
Then, the prior anchors obtained by the Algorithm \ref{alg:kmeans} are utilized as reference points, which replace the initial reference points obtained by random initialization to more quickly regress to the appropriate element shapes. 
Moreover, to further integrate the prior anchors into the network structure and reduce the inflexibility caused by fixed values, we add the MLP module after introducing the prior anchors to represent the matching process of the prior anchors. 
The visualization of random initialization and prior reference points is shown in Fig.\ref{fig:anchor_vis}. 

In summary, through Fig.\ref{fig:mapqr_and_ours}, it can be seen that we propose an enhancement method that can be applied to any network structure focusing on the design of the decoder's attention mechanism, in order to incorporate prior information into the detection of road elements.

\begin{algorithm}
  \caption{Prior Anchor Selection Algorithm}
  \begin{algorithmic}[1]
  \renewcommand{\algorithmicrequire}{\textbf{Input:}}  
  \renewcommand{\algorithmicensure}{\textbf{Output:}}  
  \label{alg:kmeans}
  \REQUIRE ~~\\ 
  Element matrix of the dataset $\mathbf{A} \in \mathbb{R}^{N \times L}$;\\
  Shape template matrix $\mathbf{U}_M \in \mathbb{R}^{N \times M}$;\\
  Number of elements $L$, number of prior anchors $N_P$,\\
  number of maximum iterations $S_{\text{max}}$, convergence threshold $\epsilon_{A}$.
  \ENSURE ~~ \\ 
  Prior anchor matrix $P_A \in \mathbb{R}^{N \times N_P}$
  \STATE \textbf{Step 1: Shape Template Transformation} 
  \STATE Compute the coefficient matrix in the shape template space:
  \[
  \mathbf{C}_A = \mathbf{U}_M^\top \mathbf{A}, \quad \mathbf{C}_A \in \mathbb{R}^{M \times L}
  \]
  \STATE \textbf{Step 2: K-Means Clustering} 
  \STATE \quad Randomly select $N_P$ samples in $\mathbf{C}_A$ as the initial cluster centers $\mathbf{C}_{\text{centers}} \in \mathbb{R}^{M \times N_P}$.
  \STATE \quad Set current iteration $s = 0$ and relative tolerance $\Delta = \infty$.
  
  \WHILE{$s < S_{\text{max}}$ \textbf{and} $\Delta > \epsilon_A$}
    \STATE \quad \textbf{Compute distance from samples to cluster centers} 
    \STATE \quad For each sample $\mathbf{c}_i \in \mathbf{C}_A$, compute Euclidean distance to all cluster centers:
    \[
    d_{i,j} = \|\mathbf{c}_i - \mathbf{C}_{\text{centers}, j}\|, \quad j = 1, \dots, N_P
    \]
    \STATE \quad Assign each $\mathbf{c}_i$ to its nearest cluster center to obtain label vector $\mathbf{L} \in \mathbb{Z}^{L}$
    \STATE \quad \textbf{Update cluster centers} 
    \STATE \quad Compute the new center of each cluster:
    \[
    \mathbf{C}_{\text{centers}, j} = \frac{1}{|\mathcal{C}_j|} \sum_{\mathbf{c}_i \in \mathcal{C}_j} \mathbf{c}_i, \quad j = 1, \dots, N_P
    \]
    where $\mathcal{C}_j$ denotes the set of data points assigned to cluster $j$.

    \STATE \quad Compute the total change in cluster centers:
    \[
    \Delta = \sum_{j=1}^{N_{P}} \|\mathbf{C}_{\text{centers}, j}^{(t+1)} - \mathbf{C}_{\text{centers}, j}^{(t)}\|
    \]
    \STATE \quad Increment iteration counter: $t = t + 1$
  \ENDWHILE

  \STATE \textbf{Step 3: Prior Anchor Selection}
  \STATE \quad For each cluster, select the coefficient value $\mathbf{c}_{\text{anchor}, j}$ closest to the cluster centers as the coefficients of prior anchor:
  \[
  P_C = \{ \mathbf{c}_{\text{anchor}, j} \}_{j=1}^{N_P}, \quad P_C \in \mathbb{R}^{M \times N_P}
  \]
  \STATE \textbf{Step 5: Inverse Transformation to Euclidean Space}
  \[
  P_A = \mathbf{U}_M P_C, \quad P_A \in \mathbb{R}^{N \times N_P}
  \]
  \STATE \textbf{Return:} Prior anchor matrix $P_A$
  \end{algorithmic}
\end{algorithm}

\begin{figure*}[h]
    \centering
    \includegraphics[width=0.8\linewidth]{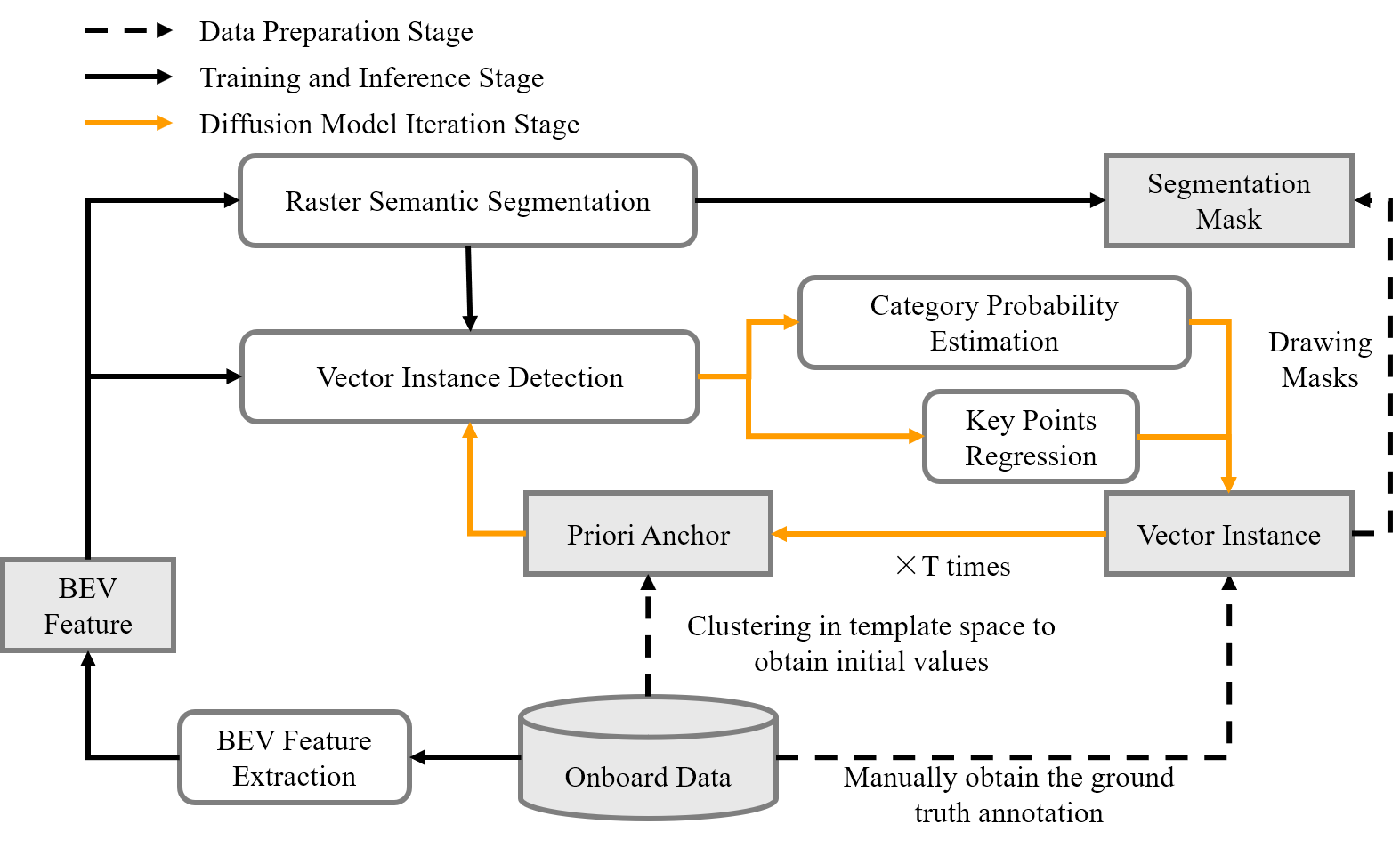}
    \caption{The overall pipeline of our vectorized road elements perception method.}
    \label{fig:architecture}
\end{figure*} 
 
\subsection{Diffusion Policy}

Based on the basic principles of the diffusion model, the forward process is defined as follows:

\begin{equation}
    q\left(\mathbf{r}^i \mid \mathbf{r}^{i-1} \right) = \mathcal{N} \left( \mathbf{r}^i ; \sqrt{\alpha^i} \mathbf{r}^{i-1}, (1 -\alpha^i) \mathbf{I} \right)
    \label{eq:chap03-eq28}
\end{equation}

Furthermore,

\begin{equation}
    q\left(\mathbf{r}^i \mid \mathbf{r}^0 \right) = \mathcal{N} \left( \mathbf{r}^i ; \sqrt{\bar{\alpha}^i} \mathbf{r}^0, (1 - \bar{\alpha}^i) \mathbf{I} \right)
    \label{eq:chap03-eq29}
\end{equation}

Here, $\mathbf{r}^0$ is the original road element instance data, and $\mathbf{r}^i$ is the data after adding noise at step $i$, following a Gaussian distribution with a mean of $\sqrt{\bar{\alpha}^i} E^0$ and a covariance of $(1 - \bar{\alpha}^i) \mathbf{I}$. The noise attenuation factor $\bar{\alpha}^i$ is defined as

\begin{equation}
    \bar{\alpha}^i  =\prod_{s=1}^{i} \alpha^s = \prod_{s=1}^{i} (1 - \beta^s)
\label{eq:chap03-eq30}
\end{equation}

where $\beta^s$ is the noise addition intensity at step $s$. This forward diffusion process can be seen as gradually perturbing the shape and distribution information of the original road space elements, thereby simulating their diverse structural forms and spatial transformations.

During the reverse diffusion inference stage, the diffusion model learns a parameterized denoising network $f_{\theta}(\mathbf{r}^i,z,i)$, which, guided by conditional information $z$, progressively refines the random noise $E^i$ sampled from a Gaussian distribution into the desired representation of road elements $\mathbf{r}^0$. The overall modeling of this process is defined in Equation \ref{eq:chap03-eq31}:

\begin{equation}
    p_{\theta}(\mathbf{r}^0 \mid z) = \int p(\mathbf{r}^T) \prod_{i=1}^{T} p_{\theta}(\mathbf{r}^{i-1} \mid \mathbf{r}^i, z) \, d{\mathbf{r}^{1:T}}
    \label{eq:chap03-eq31}
\end{equation}

Considering the high computational cost of fully executing $T$ steps, a truncated diffusion strategy is employed during training, where the denoising process is modeled only within a shorter step range. Let $N_P = N_{queries}$, indicating that all detection queries are derived from priors generated by clustering.

During training, starting from the set of prior anchors $P_A=\{{anchor}_{k}\}^{N_P}_{k=1}$ obtained through clustering, Gaussian noise is added to construct diffusion training samples:
\begin{equation}
  \mathbf{r}^i_k = \sqrt{\bar{\alpha}^i} {anchor}_{k} + \sqrt{1 - \bar{\alpha}^i} \epsilon
  \label{eq:chap03-eq32}
\end{equation}

where \( i \in [1, T_{\text{trunc}}] \) and the truncated diffusion step \( T_{\text{trunc}} \ll T \), with Gaussian noise $\epsilon \sim \mathcal{N}(0, \mathbf{I})$.

Subsequently, the conditional diffusion model is used to perform a denoising prediction on the noise-augmented anchors.

\begin{equation}
  \left\{ \hat{c}_k, \hat{\mathbf{r}}_k \right\}_{k=1}^{N_{\text{anchor}}} = f_{\theta} \left( \left\{ \mathbf{r}^i_k \right\}_{k=1}^{N_{\text{anchor}}}, z \right)
  \label{eq:chap03-eq33}
\end{equation}

where $\hat{c}_k$ denotes the predicted category of the road element and $\hat{\mathbf{r}}_k$ represents the predicted keypoint representation.

Through noise augmentation and denoising learning in the training process described above, the model effectively learns and leverages structural prior information of road space environment elements, significantly enhancing perception accuracy and shape representation capabilities.

\subsection{Overall Pipeline}

To effectively integrate shape prior information into neural network design, this study proposes a shape-prior-enhanced framework for road element perception, as illustrated in Fig. \ref{fig:architecture}. This framework is composed of the following key components.

\subsubsection{BEV Feature Extraction}
In this study, BEVformer \citep{yang2023bevformerv2} is used to get BEV features from sensor inputs in order to detect the road spatial environment under BEV space. ResNet50 is used as the backbone network to extract 2D features from sensor inputs. 
\subsubsection{Semantic Segmentation Head}
This branch generates a BEV semantic segmentation map, providing global shape and distribution information about road elements. By decoding BEV features through an up-sampling network, it offers both comprehensive environmental understanding and shape-aware prior information, guiding the subsequent vectorized instance extraction process.

\subsubsection{Shape-Prior-Guided Query Refinement Module}
This module selects a feature layer with strong shape information from the semantic segmentation head, refines it through a pooling operation, and aligns it with the vector query size. A feature interaction module is then used to transfer shape-prior information to the vector query.

\subsubsection{Shape-Prior Anchor Fusion}
We first learn shape templates of road elements from a large-scale dataset and cluster them in the template space to generate representative anchor points. These anchors, serving as "prototypes" of road element shapes, are directly fed into the Transformer instance decoder during inference, providing stable and reliable reference points for instance extraction.

\subsubsection{Vector Instance Decoder Branch}
With shape-prior-refined vector queries, we employ a multi-layer Transformer network for vector instance decoding. This branch iteratively optimizes the vector queries, progressively refining their offset predictions relative to the prior reference points. Through iterative computation, the Transformer decoder produces complete vector instances of road elements.

\subsubsection{Diffusion Model Used in Vector Decoder}
This process begins by adding noise to shape-prior anchors, which are then fed into the vector decoder branch. The decoder's output serves as the new initial reference points. After that, the loop is iterated for a finite number of T times, as described above, to obtain the final result.

\subsubsection{Training Loss}
We supervise the segmentation mask and vector instance at the end of the pipeline. During training, out model kept in line with the loss functions of other methods  \citep{liao2023maptrv2}, using category prediction loss $\mathcal{L}_{\text{cls}}$, point coordinate regression loss $\mathcal{L}_{\text{pts}}$, edge direction loss $\mathcal{L}_{\text{dir}}$, and semantic segmentation loss $\mathcal{L}_{\text{seg}}$. The difference is that we add the template space loss $\mathcal{L}_{\text{shape}}$, which transforms the predicted coordinates to the template space using the L1 loss for supervised.

    \begin{align}
        \mathcal{L}_{\text{sum}} & = \lambda_{\text{cls}} \mathcal{L}_{\text{cls}} + \lambda_{\text{pts}} \mathcal{L}_{\text{pts}} \\
   & + \lambda_{\text{dir}} \mathcal{L}_{\text{dir}} +\lambda_{\text{shape}} \mathcal{L}_{\text{shape}} +\lambda_{\text{seg}} \mathcal{L}_{\text{seg}}\label{eq:chap03-eq39}
    \end{align}
  
where $\lambda_{\text{cls}}, \lambda_{\text{pts}}, \lambda_{\text{dir}}, \lambda_{\text{seg}}, \lambda_{\text{shape}}$ are the coefficients used to adjust the weights of each task to optimize the training results.

\section{Experiments}

\subsection{Experimental Setup}

\textbf{Dataset.} Our experiments were conducted on the nuScenes dataset \citep{caesar2020nuscenes}, a widely recognized and utilized benchmark for online mapping. 
This dataset includes road data from multiple cities such as Boston and Seattle, and features a variety of scenarios with different weather conditions, road types, and traffic densities, both in daytime and nighttime settings. 
The nuScenes dataset is widely recognized as a standard for comparative evaluations. Validating our method on this dataset highlights its robustness and provides a comprehensive view of its performance across various road scenarios.

\begin{table*}[t]
  \caption{Comparison of state-of-the-art road element perception network based on camera.}
  \label{tab:chap03-tab01}
  \begin{center}
  \def\arraystretch{1.3}
  \resizebox{\textwidth}{!}{%
  \begin{tabular}{cccccccc}
      \hline
      Method & Backbone & Image Size & Epochs &$\text{AP}_{ped}$($\uparrow$) &  $\text{AP}_{div}$($\uparrow$) &  $\text{AP}_{bound}$($\uparrow$) & mAP($\uparrow$)\\
      \hline
    
      HDMapNet  \citep{li2022hdmapnet} & Effi-B0 & 480$\times$800 & 30 & 14.4\% & 21.7\% & 33.0\% & 23.0\% \\
      VectorMapNet  \citep{liu2022vectormapnet} & R50 & 480$\times$800 & 110 & 36.1\% & 47.3\% & 39.3\% & 40.9\% \\  
      MapTR  \citep{liao2022maptr} & R50 & 480$\times$800 & 24 & 46.3\% & 51.5\% & 53.1\% & 50.3\% \\
      InsightMapper  \citep{xu2023insightmapper} & R50 &480$\times$800 & 24 & 44.4\% & 53.4\% & 52.8\% & 50.2\%  \\  
      PivotNet  \citep{ding2023pivotnet} & R50 & 480$\times$800 & 30 & 53.8\% & 58.8\% & 59.6\% & 57.4\% \\
      MapTRv2  \citep{liao2022maptr} & R50 & 480$\times$800 & 24 & 59.8\% & 62.4\% & 62.4\% & 61.5\% \\
      GeMap  \citep{zhang2023onlinegemap} & R50 & 480$\times$800 & 110 & 59.8\% & 65.1\% & 63.2\% & 62.7\% \\
      MapQR(retrospective version)  \citep{liu2024mapqr} & R50 & 480$\times$800 & 24 & 61.8\% & 66.9\% & 67.0\% & 65.2\% \\
      PriorFusion-V1(Ours) & R50 & 480$\times$800 & 24 & 63.2\% & 69.3\% & 66.5\% & 66.3\% \\
      PriorFusion-V2(Ours) & R50 & 480$\times$800 & 110 & \textbf{68.3\%} & \textbf{72.0\%} & \textbf{70.9\%} & \textbf{70.4\%} \\
      
      \hline
  \end{tabular}}
  \end{center}
\end{table*}

\begin{table}[bh]
        \caption{Accuracy results of our method under the threshold $\tau=0.2$.}
        \label{tab:results02}
        \begin{center}
        \def\arraystretch{1.3}
        \begin{tabular}{ccccc}
            \hline
            Method &  $\text{AP}_{ped}$($\uparrow$) &  $\text{AP}_{div}$($\uparrow$) &  $\text{AP}_{bound}$($\uparrow$) & mAP($\uparrow$) \\
            \hline
            Baseline &  3.31\% & 19.49\% & 2.86\% & 8.55\% \\
            PriorFusion-V1 &  5.32\% & 19.58\% & 4.08\% & 9.66\% \\
            PriorFusion-V2 &  \textbf{6.84\%} & \textbf{30.82\%} & \textbf{9.77\%} & \textbf{15.81\%} \\
            
            \hline
        \end{tabular}
        \end{center}
\end{table}

\textbf{Training Details.} 
The selection of the baseline model adopted the state-of-the-art (SOTA) solution MapQR \citep{liu2024mapqr}, which achieves the highest performance metrics among works that provide reproducible code.
Moreover, as shown in Fig. \ref{fig:mapqr_and_ours}, the method proposed in this study does not affect the deep structure of the decoder, featuring a plug-and-play characteristic that allows for modular integration into other works. There is room for further improvement with alternative approaches.
The model was trained on 8 NVIDIA RTX A6000 GPUs with a batch size of 24, using the AdamW optimizer with a learning rate of $5\times10^{-4}$. The network architecture adopts ResNet50 as the backbone and leverages BEVFormer \citep{yang2023bevformerv2} for BEV feature extraction, consistent with the MapQR baseline \citep{liu2024mapqr}.

Most works represent elements using 20 sampled points with N=40. Accordingly, when constructing the shape template space, M is set to 20.
The number of diffusion steps, denoted as step num $T$, is set to 2.
The perception range is set to $[-15.0m, 15.0m]$ along the X-axis and $[-30.0m, 30.0m]$ along the Y-axis. Then the corresponding size of BEV semantic segmentation mask is $400 \times 200$.
The correlation coefficients of the training loss are set to:
$\lambda_{\text{seg}}=2$, $\lambda_{\text{cls}}=2$, $\lambda_{\text{pts}}=5$, $\lambda_{\text{dir}}=0.005$, $\lambda_{\text{shape}}=0.001$.

\textbf{Evaluation Metrics.} We evaluate road element extraction quality using the standard metrics from previous works  \citep{liao2022maptr, liu2024mapqr}. 
Specifically, we employ Average Precision (AP) as the primary evaluation metric, calculated based on Chamfer distance $D_{Chamfer}$, which determines whether a predicted element matches the ground truth (GT). The AP is defined as:
\begin{equation}
        AP  = \frac{1}{\lvert T_\theta \rvert} \sum_{\tau \in T_\theta}{AP}_\tau
\end{equation}
where ${AP}_\tau$ represents the average precision at a given Chamfer distance threshold $\tau$, where a prediction is considered a true positive only if the distance between the prediction and the GT is below the specified threshold ($\tau \in T_\theta, T_\theta=\{0.5, 1.0, 1.5\}$).

The APs of all categories were further averaged to obtain the final Mean Average Precision (mAP) metric, which is used to measure the overall detection performance of the model on all types of road elements:
\begin{equation}
  \text{mAP} = \frac{1}{N_{class}} \sum_{class=1}^{N_{class}} \text{AP}_{class}
  \label{eq:chap03-eq51}
\end{equation}
In this study, our model is consistent with the previous SOTA work in that only three classes of elements (lane dividers, road boundaries and pedestrian crosssings) are processed, thus $N_{class}=3$.

For further analysis, we also evaluate AP at a more stringent Chamfer distance threshold ($\tau = 0.2$) to highlight performance differences in capturing fine details among different methods.

To provide a more comprehensive assessment of performance in ablation study, we also measure the network's operational efficiency using Frames Per Second (FPS) and quantify the network size by counting the parameters.

\subsection{Main Results}

\begin{table*}[h]
        \caption{Ablation on the Shape-Prior-Guided (SPG) Query Refinement Module.}
        \label{tab:ablation1}
        \begin{center}
        \def\arraystretch{1.3}
        \begin{tabular}{ccccccc}
            \hline
            Method &$\text{AP}_{ped}$($\uparrow$) &  $\text{AP}_{div}$($\uparrow$) &  $\text{AP}_{bound}$($\uparrow$) & mAP($\uparrow$) & FPS & Params.\\
            \hline
            Model0 & 61.8\% & 66.9\% & 67.0\% & 65.2\% & 14.08 & 1433M\\
            Model1 & 64.4\% & 67.7\% & 66.2\% & 66.1\%& 13.54 & 1495M\\
            \hline
        \end{tabular}
        \end{center}
\end{table*}

\begin{table*}[h]
        \caption{Ablation on the Shape-Prior-Anchor.}
        \label{tab:ablation2}
        \begin{center}
        \def\arraystretch{1.3}
        \begin{tabular}{ccccccc}
            \hline
            Method &$\text{AP}_{ped}$($\uparrow$) &  $\text{AP}_{div}$($\uparrow$) &  $\text{AP}_{bound}$($\uparrow$) & mAP($\uparrow$) & FPS & Params.\\
            \hline
            Model1 & 64.4\% & 67.7\% & 66.2\% & 66.1\% & 13.54 & 1495M\\
            Model2 & 63.2\% & 69.3\% & 66.5\% & 66.3\% & 13.29 & 1504M\\
            \hline
        \end{tabular}
        \end{center}
\end{table*}

Table \ref{tab:chap03-tab01} compares the performance of our method with classical methods and state-of-the-art (SOTA) approaches, all of which rely solely on camera sensor inputs and primarily use ResNet50 (R50) as the backbone. Our method consistently outperforms the compared methods in both 24-epoch and 110-epoch training scenarios.
Specifically, PriorFusion-V1 represents our model without the diffusion step, while PriorFusion-V2 includes the diffusion step. Notably, PriorFusion-V2 was not tested for 24 epochs due to the slower convergence of the diffusion model.

Our method achieves $68.3\%$ AP for pedestrian crossings, $72.0\%$ AP for dividers, and $70.9\%$ AP for boundaries, ultimately achieving a state-of-the-art mAP of $70.4\%$ with R50 as the backbone. Even with PriorFusion-V1, our method demonstrates strong performance, particularly in the 24-epoch scenario, where it achieves an outstanding $AP_{div}$ of $69.3\%$, highlighting the superiority of our approach.

Furthermore, we evaluated AP performance at a stricter threshold of 0.2. Since this threshold is very low, the criterion for determining true positives is more stringent, providing a clearer assessment of the model's ability to capture shape regularity and completeness. The comparative results are presented in Table \ref{tab:results02}.

The results show that our model achieves significant improvements at the 0.2 threshold. In particular, PriorFusion-V1 demonstrates notable gains for pedestrian crossings and boundaries, highlighting the effectiveness of incorporating shape priors. In contrast, the improvement for dividers is less pronounced, likely because their baseline performance is already high.
In the case of PriorFusion-V2, notable improvements are observed across all categories, with the mAP increasing by 7.26\%, further demonstrating the superiority of our method.

\subsection{Ablation Study}
To validate the effectiveness of our proposed modules, we conducted ablation experiments using a replicated version of MapQR-ResNet50 \citep{liu2024mapqr} as the baseline model, trained for 24 epochs. To ensure a fair efficiency test, we measure the computational overhead on the same single Tesla A100 GPU. The calculation of parameters utilized the code provided by Maptr\citep{liao2022maptr}.

\textbf{Shape-Prior-Guided Query Refinement Module.}
Table \ref{tab:ablation1} presents the results of ablation experiments comparing the model with and without the Shape-Prior-Guided Query Refinement Module(SPG). Model0 represents the baseline, while Model1 incorporates the Shape-Prior-Guided Query Refinement Module (SPG). The results show notable improvements for pedestrian crossings and lane dividers, while the performance for boundaries slightly decreases. This decline may be attributed to the elongated nature of road edges, where mask attention is more effective for compact, well-defined elements.
Moreover, the incorporation of the SPG module has a minimal impact on computational speed and does not significantly increase the computational parameter overhead. Moreover, the incorporation of the SPG module has a minimal impact on computational speed and does not significantly increase the computational parameter overhead.

\textbf{Shape-Prior-Anchor.}
Table \ref{tab:ablation2} presents the results of ablation experiments comparing the model with and without the Shape-Prior-Anchor. Model1 is aforementioned, while Model2 extends this by incorporating the Shape-Prior-Anchor. The results indicate a substantial improvement in the dividers category, while the performance for Pedestrian crossings slightly decreases.
This performance difference can be attributed to the characteristics of the shape-prior anchor. Since the anchor includes more dividers and fewer pedestrian crossings, and because dividers have simpler, more consistent shapes, the model benefits significantly in this category. In contrast, the complex shape and variable distribution of pedestrian crossings lead to constrained predictions when using the shape-prior anchor, resulting in a slight performance decline.
The introduction of Shape-Prior-Anchor results in a decrease of 0.25 FPS and an increase of 9M in parameters, both of which are due to the newly added MLP module shown in Fig.\ref{fig:mapqr_and_ours} (b).

\textbf{Diffusion Steps.}
Table \ref{tab:ablation3} presents the comparison between Model3 (without the diffusion step) and Model4 (with the diffusion step). Due to the slower convergence of the diffusion model, both models were trained for 110 epochs to ensure a fair comparison. The results clearly show that incorporating the diffusion step leads to consistent improvements across all metrics, demonstrating the effectiveness of this approach.

\begin{table*}[h]
        \caption{Ablation on the Diffusion Model.}
        \label{tab:ablation3}
        \begin{center}
        \def\arraystretch{1.3}
        \begin{tabular}{ccccccc}
            \hline
            Method &$\text{AP}_{ped}$($\uparrow$) &  $\text{AP}_{div}$($\uparrow$) &  $\text{AP}_{bound}$($\uparrow$) & mAP($\uparrow$) & FPS & Param.\\
            \hline
            Model3 & 67.0\% & 71.2\% & 69.6\% & 69.3\%& 13.29 & 1504M\\
            Model4 & 68.3\% & 72.0\% & 70.9\% & 70.4\%
            & 7.98 & 1501M\\
            \hline
        \end{tabular}
        \end{center}
\end{table*}

To further investigate the runtime overhead introduced by the diffusion module, we evaluate the elapsed time per frame (in milliseconds) under different diffusion step settings on a single Tesla A100 GPU. Specifically, we compare the runtime of Model 3 (without the diffusion module) against Model 4 configurations where the diffusion step number is set to 1, 2, and 3. We also compute the time increment between consecutive configurations to isolate the additional cost introduced by each diffusion step.

As shown in Fig.\ref{fig:architecture}, the decoder runs in a loop according to the number of diffusion step num $T$.
For instance, when $T=2$, the truncated diffusion loop executes twice, and accordingly, the vector decoder is also used twice. Similarly, with $T=3$, the decoder runs three times in the diffusion loop.

As shown in Table.\ref{tab:ablationtime}, setting $T=1$ introduces an additional 3.11 ms of latency compared to the baseline without diffusion. This initial overhead is likely due to the inclusion of related libraries, variables, and code branching required to activate the diffusion module.

From the observed time increments between $T=1,2,3$, we estimate that each additional decoder execution incurs an average of 46.89 ms of runtime. This indicates that the primary computational overhead introduced by the diffusion mechanism is attributable to the repeated execution of the decoder, and scales linearly with the number of diffusion steps.

Although the introduction of the diffusion denoising loop increases the runtime, our model achieves a favorable trade-off between computational accuracy and efficiency at \( T = 2 \). Compared to other diffusion mapping works, the runtime remains acceptable.
\begin{table}[h]
        \caption{Ablation on the Diffusion steps.}
        \label{tab:ablationtime}
        \begin{center}
        \def\arraystretch{1.3}
        \begin{tabular}{ccc}
            \hline
            Step Num & Elapsed Time(ms) & Time Increment(ms)\\
            \hline
            No Diffusion &  75.23 & - \\
            1 &  78.34 & 3.11  \\
            2 &  125.24 & 46.90\\
            3 &  172.12 & 46.88\\
            \hline
        \end{tabular}
        \end{center}
\end{table}
\textbf{Summary.}
In summary, through the ablation study, we have verified the effectiveness of each module. By comparing computational speed and parameters, we demonstrate that the primary computational overhead and parameter increase are still concentrated in the decoder network. This confirms that our method can be easily transferred to other works with minimal overhead while providing significant accuracy gains.

\subsection{Visualization}
\begin{figure*}[p]
    \centering
    \includegraphics[width=1\linewidth]{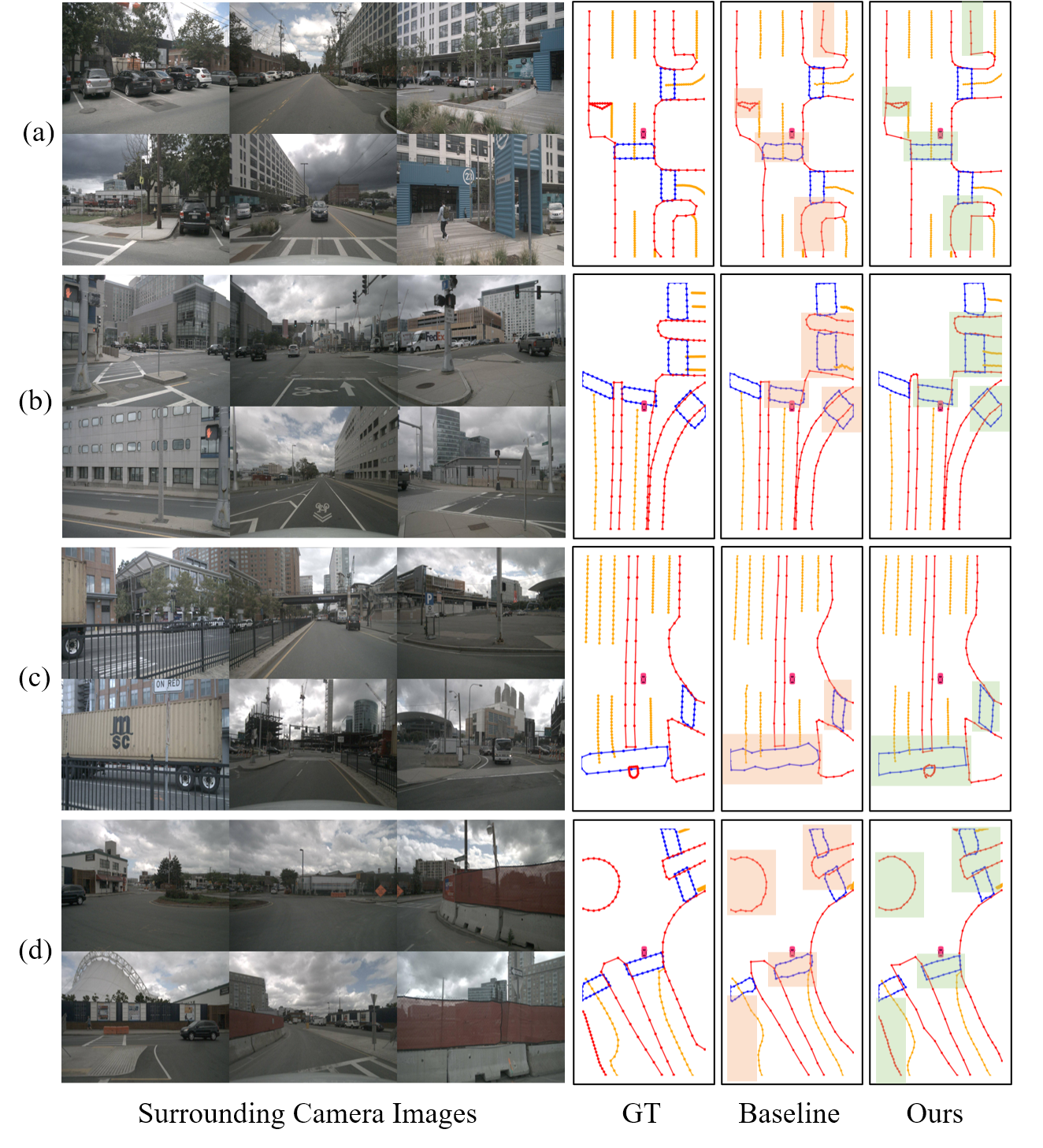}
    \caption{Visualization results. The visualized scene images are difficult scenes selected by the baseline method. This figure compares the visualization results of this study, the baseline method, and ground truth. Among them, red represents the road boundaries, yellow represents lane dividers, blue represents pedestrian crossings, orange area marks the irregular or missed shape predicted by the baseline method, and green area is the corresponding area of our method.}
    \label{fig:result2}
\end{figure*}
\begin{figure*}[p]
    \centering
    \includegraphics[width=1\linewidth]{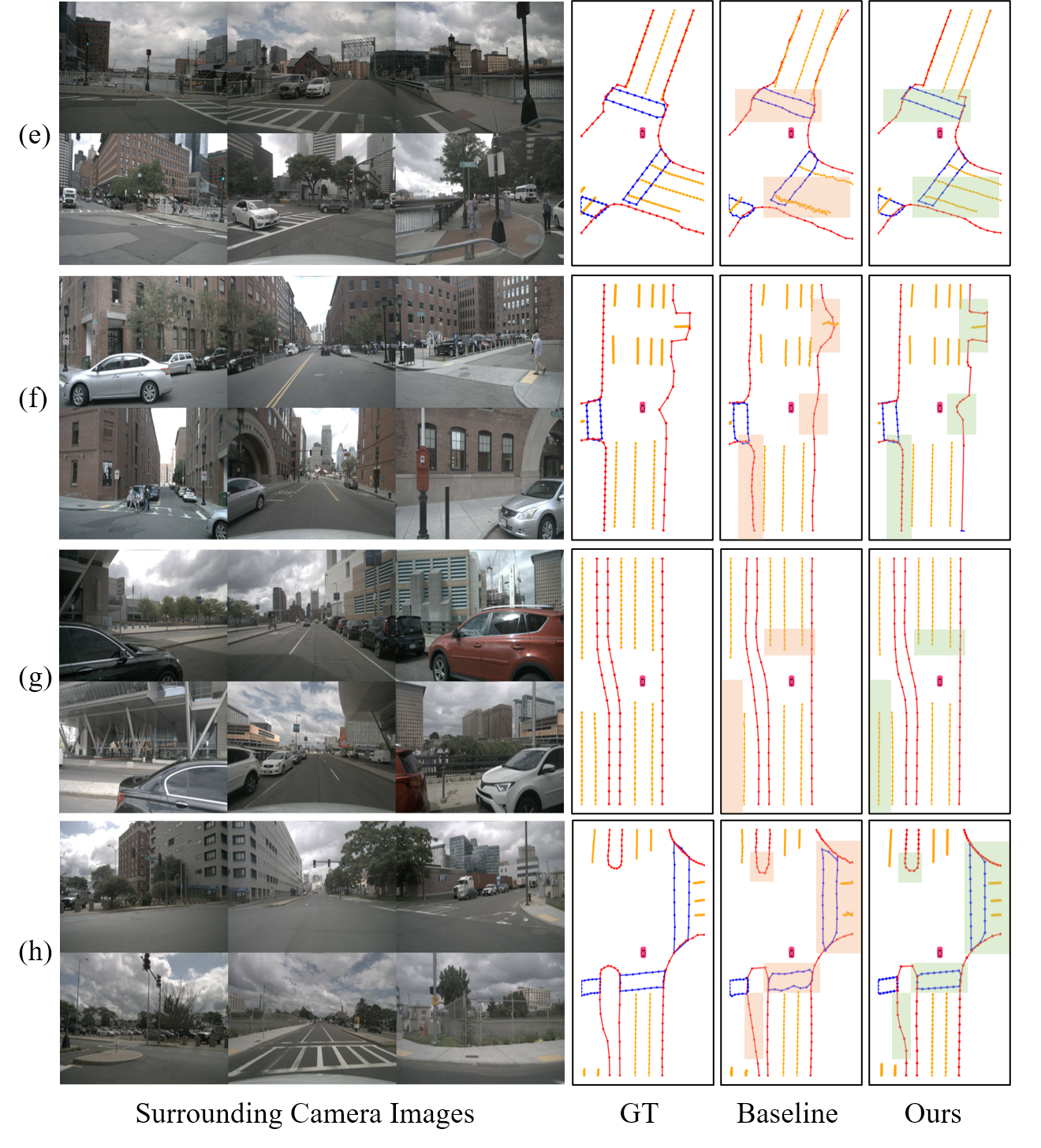}
    \caption{Additional Visualization results.}
    \label{fig:result}
\end{figure*}

 
Fig. \ref{fig:result} presents the visualization results of our method on challenging scenes selected by the widely-used baseline model MapTRv2 \cite{liao2023maptrv2}. 
These scenarios are commonly used for visual comparison in related literature. To the best of our knowledge, our method produces predictions that most closely match the ground truth in terms of edge smoothness, shape regularity, and clarity at critical turning points.

In Fig. \ref{fig:result} (a), our model’s predicted edge contours align more closely with shape rules, exhibiting smooth, regular boundaries without irregular fluctuations. The top-marked boundary is also predicted with greater linearity.

In Fig. \ref{fig:result} (b), our method accurately predicts pedestrian crossings with clearer, well-defined edges, while the baseline method fails to detect one divider—an issue our approach resolves by leveraging prior distribution information.

In Fig. \ref{fig:result} (c), the baseline model produces highly distorted shapes for the pedestrian crossing and fails to detect a subtle but critical road boundary. Such omissions can mislead downstream decision-making modules by misidentifying drivable and hazardous regions. In contrast, our model provides stable and accurate predictions that better reflect the actual road layout.

In Fig. \ref{fig:result} (d), the upper right region of the image shows that our method better adheres to intersection drawing rules. Moreover, our method accurately captures the roundabout boundary on the left with a more rounded shape. In contrast, the baseline method misses detecting the curb in the lower left corner, a gap effectively addressed by our approach.

Fig.\ref{fig:result2} provides more visual comparisons, further demonstrating the superiority of our method in terms of both shape fidelity and spatial distribution. 

Overall, compared to both the baseline predictions and ground truth (GT), our method consistently produces more regular shapes, more complete element distributions, and significantly fewer missed detections.

\section{Discussion and Conclusion}

In this work, we proposed PriorFusion, a shape-prior-enhanced perception framework that integrates semantic, geometric, and generative priors into a unified architecture. By incorporating a shape-prior-guided query refinement module, clustering in a shape template space, and a truncated diffusion mechanism, our method significantly improves the accuracy and geometric regularity of road element detection.

Extensive experiments on the nuScenes dataset demonstrate that PriorFusion consistently outperforms classical and state-of-the-art baselines. Notably, our method achieves a 7.26\% mAP gain under a stringent evaluation threshold of $\tau = 0.2$, indicating superior fine-grained alignment with ground truth and improved shape fidelity. To the best of our knowledge, in typical complex scenarios curated by MapTR \citep{liao2022maptr}, our model exhibits the smoothest edge contours and the clearest turning points in visualization, highlighting the benefits of incorporating shape priors across the pipeline.

Beyond accuracy, PriorFusion is designed with practical deployment. The truncated diffusion strategy introduces minimal step number, achieving 7.98 FPS with just two decoder iterations. Additionally, ablation studies show that the inclusion of the shape-prior-guided (SPG) module and anchor priors imposes negligible overhead in both runtime and parameter count. This efficiency makes PriorFusion readily deployable as a \textbf{plug-and-play module} in other road perception systems to enhance both robustness and geometric consistency.

\subsection{Limitations and Discussion on Prior Knowledge}
Despite the demonstrated improvements, our method has several limitations. 

Although prior anchors are effective in capturing dominant shape structures, they may not generalize well to drastically different road layouts or urban designs, limiting cross-scenario adaptability. 
The PriorFusion relies on dataset-specific clustering to construct the shape template space, requiring re-clustering when applied across datasets. While the use of SVD enables effective dimensionality reduction, clustering over large-scale data still incurs non-negligible memory costs. 

In terms of coping with nonlinear complex shapes, although the template space can be expressed to some extent, errors still exist. We made preliminary attempts to apply non-linear alternatives such as VAEs for shape embedding, but they were not successfully implemented in our current framework.

We recognize the inherent risks of relying on static prior knowledge. While our framework introduces priors through data-driven learning, semantic segmentation, and generative modeling—instead of directly imposing fixed HD maps—over-reliance on priors can lead to inflexible detection or false positives when the current scene deviates significantly from learned distributions. For example, incomplete observations of road geometry due to temporary occlusions may conflict significantly with the prior shape, leading to erroneous judgments. This highlights the importance of employing more robust prior models such as vision-language models (VLMs).
\subsection{Future Work}
We plan to explore several directions to further enhance the adaptability and efficiency of our framework. These include the development of cross-scenario transferable priors, such as anchor adaptation algorithms tailored for new domains; the adoption of more powerful non-linear dimensionality reduction techniques, including improved VAE-based embeddings, to better capture complex and diverse shape variations; and the integration of diffusion models with VLMs to enable semantic-aware online mapping. In addition, we aim to further optimize diffusion-based decoding to achieve real-time performance even on resource-constrained platforms.
\subsection{Conclusion}
PriorFusion provides a principled and scalable solution for accurate and geometry-aware road element perception, without relying on high-definition maps. It paves the way for more structured, interpretable, and deployable perception systems in autonomous driving. Future work will focus on addressing the limitations identified and exploring new directions to enhance the robustness and efficiency of our framework.

\section*{Acknowledgements}
This work was supported in part by National Natural Science Foundation of China (52472449, U22A20104, 52394264, 52402499), in part by Beijing Natural Science Foundation (L231008, 23L10038), in part by China Postdoctoral Science Foundation (2024M761636), the State-funded postdoctoral researcher program of China (GZC20231285), in part by the Tsinghua University-Toyota Joint Center, and in part by Tsinghua University-Didi Joint Research Center.

\appendix

\bibliographystyle{elsarticle-harv} 
\bibliography{example}






\end{document}